\documentclass[]{spie}  

\usepackage{amsmath,amsfonts,amssymb}
\usepackage{graphicx}
\usepackage[colorlinks=true, allcolors=blue]{hyperref}

\usepackage[linesnumbered,ruled,vlined]{algorithm2e}
\usepackage{sidecap}
\DontPrintSemicolon

\usepackage{todonotes}

\title{Convex Potential Learning for Optimal Transport Between Medical Images}

\author[a]{Hongyu Kan}
\author[c]{Ivan Medri}
\author[b]{Natasha Ironside}
\author[c]{Kristofor Pas}
\author[c]{Richard Corp}
\author[d]{Shinjini Kundu }
\author[a,c]{Gustavo Kunde Rohde}
\affil[a]{George Washington University, Washington, DC 20052, USA}
\affil[b]{University of Illinois Chicago, Chicago, IL 60607, USA}
\affil[c]{University of Virginia, Charlottesville, VA 22903, USA}
\affil[d]{Washington University in St. Louis, St. Louis, MO 63110, USA}

\authorinfo{Further author information: (Send correspondence to Gustavo Kunde Rohde)\\Gustavo Kunde Rohde: E-mail: gustavo.rohde@gmail.com}

\begin{document} 
\maketitle

\begin{abstract}

Computing optimal transport maps between medical images plays a fundamental role in morphological analysis. However, existing methods are either difficult to scale to large 3D medical images or rely on regularized formulations that provide only approximate transport maps. To improve the accuracy of downstream medical image analysis, we propose a novel approach for estimating optimal transport maps by parameterizing the convex potential function. By construction, the resulting transport map is guaranteed to be curl-free, consistent with Brenier's theorem. We evaluate the proposed method on both 2D and 3D medical imaging datasets. Experimental results demonstrate that our approach achieves lower transport estimation errors and produces higher-quality image reconstructions than existing methods.
 
\end{abstract}


\section{Introduction}
\label{sec:intro}  

Optimal transport has found a wide range of applications in medical image analysis, including image registration\cite{haker2004optimal}, modeling biological dynamics\cite{ozturk2023continuous}, cellular morphometry\cite{basu2014detecting}, and tissue and organ morphometry\cite{ironside2025predictive}. Many of these applications require the computation of accurate optimal transport maps between images\cite{kundu2024discovering,kundu2018discovery,ironside2025predictive}. However, medical images are typically two-dimensional (2D) or three-dimensional (3D) images comprising millions of pixels or voxels, which makes the computation of optimal transport maps particularly challenging. Existing mainstream methods are either limited to small-scale transport problems or rely on regularized formulations that produce only approximate solutions\cite{flamary2021pot,kundu2018discovery}.The resulting approximation errors can significantly affect downstream analyzes, particularly in transport-based morphometry\cite{kundu2018discovery,kan20263d}, where accurate transport maps are critical for feature extraction, statistical modeling, and visualization. Consequently, developing algorithms capable of accurately computing optimal transport maps for large-scale 3D medical images remains an important challenge.


\begin{figure}[!h]
    \centering
    \includegraphics[width=0.6\linewidth]{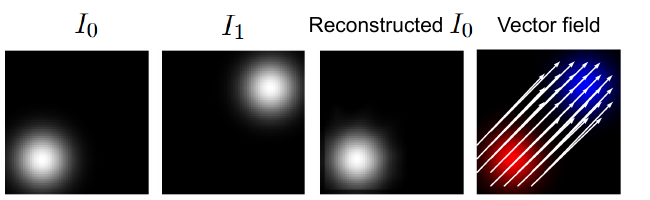}
    \caption{An example of optimal transport between two Gaussian distributions with different centers. $I_0$ (left) and $I_1$ (second from left) denote the source and target distributions, respectively. The estimated optimal transport map pushes the mass of $I_0$ onto $I_1$. The reconstructed source image (second from the right) obtained via formula \eqref{eq:recon}  and  the estimated transport map $f$, closely matches $I_0$, demonstrating accurate push-forward consistency. The rightmost panel shows the estimated transport vector field, where the red and blue contours represent $I_0$ and $I_1$, respectively, and the white arrows indicate the direction of the optimal transport map computed by the proposed method.
}
    \label{fig:syn}
\end{figure}


In this paper, we propose a novel method for estimating optimal transport maps between medical images by parameterizing and optimizing a convex potential function. The gradient of the learned convex potential function then by construction curl free, and by Brenier's theorem is an optimal transport map. If such a map can be found between two images, it is the guaranteed to be an optimal transport map between them. The proposed method is evaluated on both synthetic data and real-world 2D and 3D medical imaging datasets. The synthetic example is shown in Figure \ref{fig:syn}. Extensive experiments demonstrate that our method consistently outperforms existing approaches\cite{kundu2018discovery} in terms of push-forward matching quality, yielding more accurate transport maps and better alignment between the transported source distribution and the target distribution.


\section{Method}

\subsection{Optimal Transport Formulation for Medical Images}

Let $\Omega$ be a subset of $\mathbb{R}^d$, and $I_0$ and $I_1$ two  probability density functions over $\Omega$.
In this setting, $I_0$ and $I_1$ represent medical images to be compared. 
By nature, most medical imaging modalities, such as MRI, produce nonnegative image intensities. CT images, on the other hand, can be preprocessed to obtain nonnegative intensities. These intensities can then be normalized to sum to one and modeled as probability densities.
A mapping $f:\Omega \to \Omega $ is said to be mass-preserving if it satisfies
\begin{equation}
\label{eq:recon}
    det(Df(\textbf{x}))
    I_1(f(\textbf{x})) = I_0(\textbf{x}) , \ \forall \textbf{x} \in \Omega
\end{equation}
where $Df(\textbf{x})$ denotes the Jacobian matrix of $f$ at $\textbf{x}$. Let $MP:=\{f:\Omega \to \Omega |  det(Df(\textbf{x}))
I_1(f(\textbf{x})) = I_0(\textbf{x}) , \ \forall \textbf{x} \in \Omega \}$ denote the set of all mass-preserving mappings. The optimal transport problem seeks the mass-preserving map that minimizes the total transportation cost:
\begin{equation}\label{eq:monge_map}
    \begin{aligned}
    f^*(\textbf{x}) = \arg \min_{f\in MP} \int_{\Omega} c(\textbf{x},f(\textbf{x}))I_0(\textbf{x})d\textbf{x}
    \end{aligned}
\end{equation}
where $c: \Omega \times \Omega \to R^+$ is the cost function.

In this work, we model medical images as density functions sampled from a continuous domain $\Omega \subseteq R^d$. Let $\Omega_d$ represent a regular grid sampling the domain $\Omega$ of the continuous image. We then adopt the squared Euclidean distance $c(\textbf{x},\textbf{y}) = \frac{1}{2} |\textbf{x}-\textbf{y}|^2$ as the transportation cost. Under this formulation, the optimal transport map between two medical images $I_0$ and $I_1$ is defined as:
\begin{equation}\label{eq:monge_map}
    \begin{aligned}
    f^*(\textbf{x}) = \arg \min_{f\in MP} \sum_{\mathbf{x} \in \Omega_d } \frac{1}{2} |\textbf{x}-f(\textbf{x})|^2I_0(\textbf{x}).
    \end{aligned}
\end{equation}
Finally, as grayscale digital image values $I(\mathbf{x})$ are only given for $\mathbf{x} \in \Omega_d$ (discrete, sampled coordinates), we adopt a trilinear interpolation scheme to compute the value of the image at arbitrary values $I(f(\mathbf{x})$.

\subsection{Brenier Map Estimation}

According to Brenier's theorem\cite{brenier1991polar}, when the transportation cost is the squared Euclidean distance, the optimal transport map $f^*$ from $I_0$ to $I_1$ is given by the gradient of a convex potential function. Specifically under suitable regularity assumptions, there exists a unique convex function $F:\mathbb{R}^d\to\mathbb{R}$ such that $f:=\nabla F:\Omega \to \Omega$ satisfies the mass-preserving constraint 
$\det(Df(\mathbf{x})),I_1(f(\mathbf{x}))=I_0(\mathbf{x})$, and the optimal transport map coincides exactly with this $f$. 

Consequently, the calculation of the optimal transport map can be reduced to the estimation of this uniquely determined convex potential function $F$. Once $F$ is obtained, the optimal transport map in \eqref{eq:monge_map} is given by $f^* = \nabla F$. However, directly searching over the space of all convex functions is impractical due to its infinite-dimensional nature. To make the search over convex functions computationally feasible, we introduce a parameterized family of convex functions $F_\theta$ and optimize the parameters $\theta$ to approximate the optimal potential $F^*$. In this work, we adopt the Log-Sum-Exp (LSE) representation:
\begin{equation}\label{eq:lse}
    F_{A,b, \beta}(\textbf{x}) = \frac{1}{\beta} log(\sum_i e^{\beta(A_i^T\textbf{x}+b_i)})
\end{equation}
where $\textbf{x} \in R^d$ is an independent variable, $\beta > 0$, $A \in R^{n\times d}$ and $b \in R^n$ are parameters. The LSE representation is convex in $\textbf{x}$, provides smooth gradients, and allows the smoothness of the potential to be controlled through the parameter $\beta$. Consequently, \eqref{eq:monge_map} can be reformulated as:
\begin{equation}\label{eq:convex}
\begin{aligned}
    F_{A,b,\beta}^* = \arg \min_{F} L(F_{A,b,\beta},I_0, I_1) :&= \sum_{\textbf{x}} \frac{1}{2}((det(\nabla^2 F_{A,b,\beta}(\textbf{x}))I_1(\nabla F_{A,b,\beta}(\textbf{x}))-I_0(\textbf{x}))^2 \\
    \end{aligned}
\end{equation}

As $ L(F_{A,b,\beta},I_0, I_1)$ approaches zero, $F_{A,b,\beta}$ converges to an approximation of $F^*$. By Brenier's theorem, the corresponding transport map is then approximated by $\nabla F_{A,b,\beta}$, yielding an approximate optimal transport map between $I_0$ and $I_1$.

\subsection{Algorithm}
We optimize the parameters of $F_{A,b,\beta}$ using gradient descent by minimizing the loss function $L(F_{A,b,\beta}, I_0, I_1)$. PyTorch provides an automatic differentiation framework that allows us to efficiently compute the gradients of the loss with respect to the model parameters and perform gradient-based updates. However, computing $L(F_{A,b,\beta},I_0, I_1)$ requires evaluating the mass-preserving error over all pixels (or voxels). For large-scale medical images, this can involve millions of spatial locations, most of which correspond to zero-valued background and contribute negligibly to the optimization. Inspired by the Top-k loss\cite{lapin2016loss}, we therefore evaluate the loss at each iteration using only the pixels with the largest errors. This hard-example mining strategy focuses the optimization on the most informative regions, substantially reducing the computational cost while preserving optimization performance.

During our experiments, we found that jointly optimizing the smoothness parameter $\beta$ often led to unstable convergence and degraded optimization performance. Therefore, rather than updating $\beta$ through gradient descent, we fix it and optimize only the parameters $A$ and $b$.
The optimization procedure is summarized in Algorithm \eqref{alg}.

\begin{algorithm}[!h]
\label{alg}
  \caption{Convex Potential}
  \KwIn{$I_0$, $I_1$, $\beta$, $\eta$, $\epsilon$, $k$}
  \KwOut{$f$}
  \BlankLine       
  Random initialize parameters $A, b$ \;
  \Repeat{$L(F_{A,b,\beta},I_0, I_1) < \epsilon$}{
    Compute per-pixel errors $e(\textbf{x}) = \frac{1}{2}((det(\nabla^2 F_{A,b,\beta}(\textbf{x}))I_1(\nabla F_{A,b,\beta}(\textbf{x}))-I_0(\textbf{x}))^2$ \;
    $\Omega_k \gets $indices of the $k$ largest values in ${e(\textbf{x})}$ \;
    Compute $L_{topk} = \sum_{\textbf{x}\in \Omega_k} e(\textbf{x})$ \;
    $A, b \gets A , b - \eta \nabla_{A,b} L_{topk} $ \;
  }
  $f \gets \nabla F_{A,b,\beta}$ \;
  \Return $f$ \;
\end{algorithm}

\section{Experimental results}

\subsection{Implementation details}

We optimize \eqref{eq:convex} using gradient-based optimization. Specifically, the parameters of $F_{A,b,\beta}$ are first initialized, after which the gradients are automatically calculated using PyTorch and the parameters are updated with the Adam optimizer. We set $\beta=30$ and the initial learning rate at $0.01$. A multi-step learning rate schedule is adopted. Specifically, whenever optimization reaches a plateau, the learning rate is reduced by a factor of $0.1$. The optimization ends when the learning rate decreases below $10^{-4}$ or the loss converges. For the Top-k loss, we set
$k=\min(10^{6},|\Omega|)$
where $|\Omega|$ denotes the total number of pixels in the image. All experiments were conducted with PyTorch 2.11.0 on a single NVIDIA RTX 5000 or NVIDIA A100 GPU. For the compared method\cite{kundu2018discovery}, we followed their implementation, including image smoothing and multi-scale optimization techniques.



\subsection{Dataset}

We evaluated the accuracy of the proposed optimal transport maps on both a 2D cellular dataset\cite{basu2014detecting} and a 3D brain hematoma dataset\cite{ironside2025predictive}. The 2D cellular dataset is a nuclei dataset consisting of 80 cell images, each with a resolution of 64 × 64 pixels\cite{wang2010optimal}. The 3D dataset consists of 10 CT volumes of lobar intracerebral hemorrhage (ICH). Each CT volume has a spatial size of 150 × 190 × 120 voxels. 
Before the experiments, all images were preprocessed by segmenting the target structures, followed by centering and intensity normalization\cite{kan20263d}. We compute the optimal transport map for all ordered image pairs in the dataset, excluding self-pairs.

\subsection{Matching Comparison}


According to \eqref{eq:monge_map}, the optimal transport map satisfies the constraint on mass preservation $det(Df(\textbf{x})) I_1(f(\textbf{x})) = I_0(\textbf{x})$. Therefore, for an estimated transport map, the discrepancy between $det(Df(\textbf{x})) I_1(f(\textbf{x}))$ and $I_0(\textbf{x})$ provides a natural measure of its precision. We evaluate this discrepancy using both the $L_1$ \eqref{eq:l1} and the $L_2$ \eqref{eq:l2} errors for each image on the two datasets and compare the proposed method with existing approaches\cite{kundu2018discovery}. 
\begin{equation}
\label{eq:l1}
    L_1=\sum_{\textbf{x}}|det(Df(\textbf{x})) I_1(f(\textbf{x})) - I_0(\textbf{x})|d\textbf{x}
\end{equation}
\begin{equation}
    \label{eq:l2}
    L_2=\sum_{\textbf{x}}(det(Df(\textbf{x})) I_1(f(\textbf{x})) - I_0(\textbf{x}))^2d\textbf{x}
\end{equation}

\begin{table}[!h]
    \centering
    \begin{tabular}{c|cc|cc}
         & \multicolumn{2}{c|}{Cell dataset reconstruction error} & \multicolumn{2}{c}{ICH dataset recontruction error}  \\
        Method & $L_1$ & $L_2$  & $L_1$ & $L_2$\\
        \hline
         S. Kundu et al. \cite{kundu2018discovery} & 0.334 ± 0.187 & (21.04 ± 16.06) $*10^{-5}$& 1.098 ± 0.398& (7.89 ± 6.24) $*10^{-5}$ \\
         Convex Potential & 0.239 ± 0.068 & (13.74 ± 7.88) $*10^{-5}$& 0.287 ± 0.139& (1.31 ± 1.72) $*10^{-5}$
    \end{tabular}
    \caption{$L_1$ and $L_2$ distances comparison}
    \label{table:norm}
\end{table}

\begin{figure}[!h]
    \centering
    \includegraphics[width=0.95\linewidth]{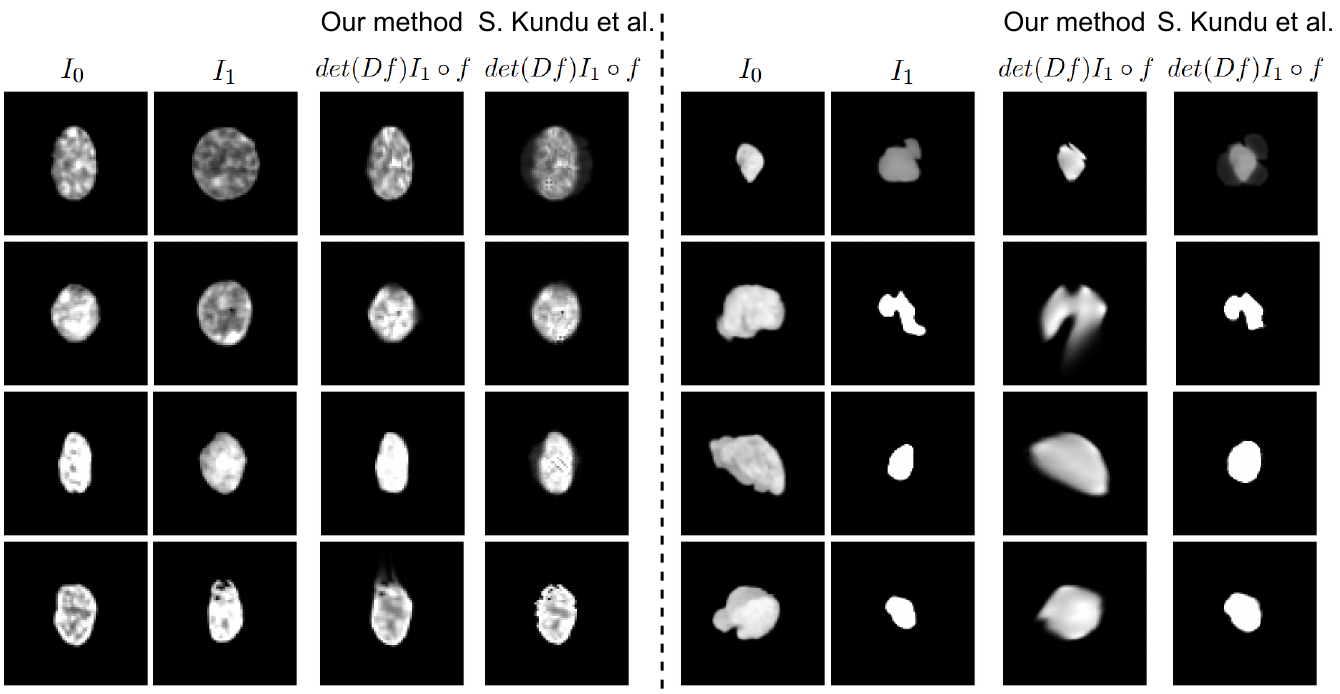}
    \caption{Visual comparison of image reconstruction using the estimated transport maps. The reconstructed images are compared with the reference image $I_0$ according to $det(Df(\textbf{x})) I_1(f(\textbf{x}))$. The left panel shows representative samples from the cell dataset, while the right panel presents the middle slice of representative samples from the ICH dataset.}
    \label{fig:exp}
\end{figure}

The quantitative results are reported in Table \ref{table:norm}. The proposed method significantly outperforms competing methods\cite{kundu2018discovery} with respect to both $L_1$ and $L_2$ error metrics.
Figure \ref{fig:exp} presents representative visualization results on the two datasets. The reconstructed images obtained using the proposed method more closely match the reference image $I_0$, indicating that the estimated transport maps more accurately push the mass distribution from $I_0$ to $I_1$. Furthermore, careful inspection reveals that the reconstructed images preserve some topological characteristics of $I_1$, such as concavities, convexities, and connected components. This behavior is expected because the transport map is computed through a smooth mapping whose Jacobian determinant remains positive, thereby preventing local folding and preserving the geometric structure of the deformation\cite{christensen1996deformable, figalli2010regularity}.

\begin{figure}[!h]
    \centering
    \includegraphics[width=1.0\linewidth]{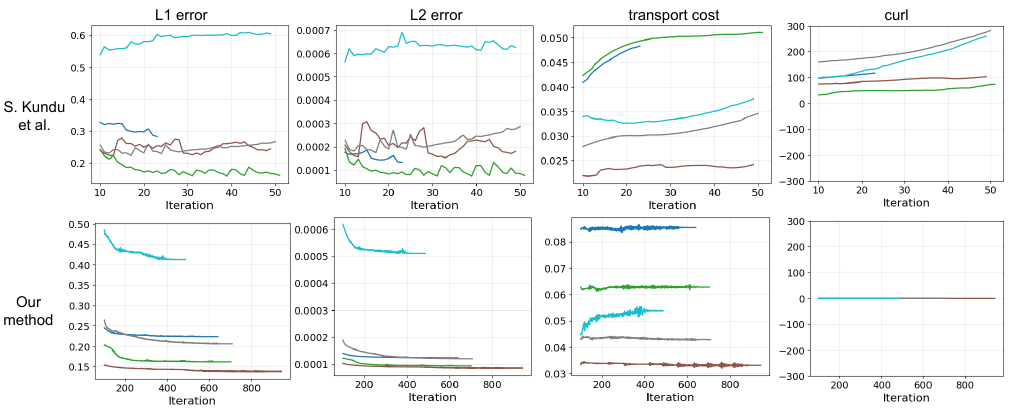}
    \caption{Comparison of the evolution of the $L_1$ error, $L_2$ error, transport cost, and curl over optimization iterations in cell dataset. Curves of the same color correspond to the same image pair.}
    \label{fig:cmp}
\end{figure}

\begin{figure}[!h]
    \centering
    \includegraphics[width=1.0\linewidth]{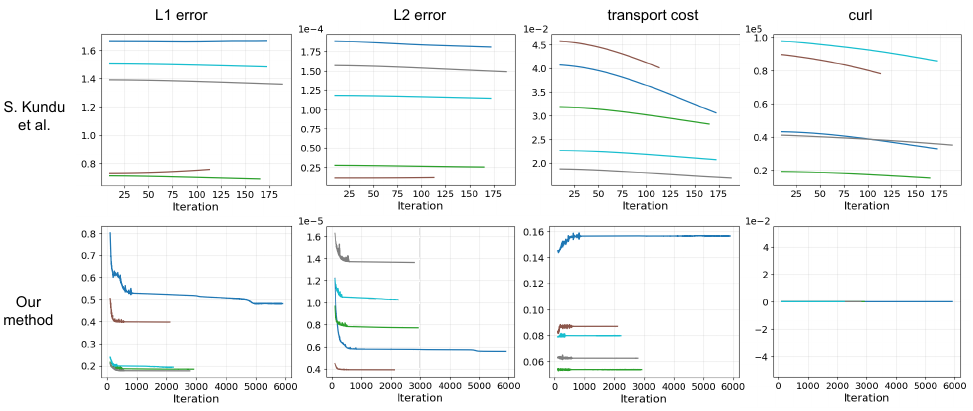}
    \caption{Comparison of the evolution of the $L_1$ error, $L_2$ error, transport cost, and curl over optimization iterations in ICH dataset. Curves of the same color correspond to the same image pair.}
    \label{fig:cmp3d}
\end{figure}

Figure \ref{fig:cmp} compares the evolution of $L_1$ error, $L_2$ error, transport cost, and curl over optimization iterations for five pairs of 2D images. Each curve corresponds to the optimization process for one pair of images. As shown in the figure, the proposed method consistently converges to lower $L_1$ and $L_2$ errors than the method of S. Kundu et al.\cite{kundu2018discovery}, indicating a more accurate push-forward matching between the transported source distribution and the target distribution. Moreover, due to the gradient-of-convex-function parameterization adopted by the proposed method, the estimated transport map is curl-free throughout the optimization process, resulting in a curl value of zero. Although the method of S. Kundu et al. achieves a lower transport cost, this is because its estimated transport map does not accurately push the source mass onto the target distribution, leading to inferior push-forward matching despite the lower transport cost.

Figure \ref{fig:cmp3d} compares the evolution of the $L_1$ and $L_2$ error, transportation cost, and curl over the optimization iterations for five representative pairs of 3D images. Similar to the observations on the 2D dataset, the proposed method consistently converges to lower $L_1$ and $L_2$ errors while maintaining a near-zero curl, demonstrating its superior accuracy and consistency in estimating optimal transport maps.




\section{Conclusion}

In this paper, we propose a novel method for estimating optimal transport maps between medical images by parameterizing and optimizing a Log-Sum-Exp (LSE) potential function. The gradient of the learned function is used to approximate the optimal transport map. We evaluate the proposed method on two medical imaging datasets, where it consistently outperforms existing approaches. Future work will consist of integrating this new method for estimating optimal transport maps into the Linear Optimal transport approach\cite{kolouri2016continuous}, and transport-based morphometry\cite{kundu2018discovery,kan20263d}, and applying it to important problems such as analyzing gray and white matter distributions\cite{kundu2024discovering}, hemorrhage expansion\cite{ironside2025predictive} and others\cite{lopes2024cell}. 



\clearpage

\acknowledgments 

Authors gratefully acknowledge funding from the ONR (N000142212505) and the NIH
(GM130825) in contributing to a portion of this work.

\bibliography{report} 
\bibliographystyle{spiebib} 

\end{document}